\documentclass[10pt,twocolumn,letterpaper]{article}
\usepackage{enumitem}

\usepackage[pagenumbers]{cvpr} %

\usepackage{graphicx}
\usepackage{amsmath}
\usepackage{amssymb}
\usepackage{booktabs}
\usepackage{algorithm}
\usepackage{algorithmic}
\usepackage[dvipsnames]{xcolor}
\usepackage{color}
\usepackage[pagebackref,breaklinks,colorlinks]{hyperref}

\usepackage[capitalize]{cleveref}
\crefname{section}{Sec.}{Secs.}
\Crefname{section}{Section}{Sections}
\Crefname{table}{Table}{Tables}
\crefname{table}{Tab.}{Tabs.}

\def\cvprPaperID{10425} %
\def\confName{CVPR}
\def\confYear{2023}

\usepackage{amsmath}

\renewcommand{\paragraph}[1]{\vspace{0.1cm}\noindent\textbf{#1}}

\begin{document}

\title{Multi-Object Manipulation via Object-Centric Neural Scattering Functions}

\author{Stephen Tian\textsuperscript{1}\thanks{indicates equal contribution. Yancheng is affiliated with Fudan University; this work was done while he was a summer intern at Stanford.} \hspace{5mm}
Yancheng Cai\textsuperscript{1}\footnotemark[1]  
\hspace{5mm} 
Hong-Xing Yu\textsuperscript{1} \hspace{5mm}
Sergey Zakharov \textsuperscript{2} \hspace{5mm}
\vspace{0.15cm}\\ 
Katherine Liu\textsuperscript{2} \hspace{5mm}
Adrien Gaidon\textsuperscript{2} \hspace{5mm}
Yunzhu Li \textsuperscript{1} \hspace{5mm}
Jiajun Wu\textsuperscript{1} \hspace{5mm}
\vspace{0.25cm}\\
\textsuperscript{1}Stanford University \hspace{10mm}
\textsuperscript{2} Toyota Research Institute
}

\maketitle
\begin{abstract}
Learned visual dynamics models have proven effective for robotic manipulation tasks. Yet, it remains unclear how best to represent scenes involving multi-object interactions. Current methods decompose a scene into discrete objects, but they struggle with precise modeling and manipulation amid challenging lighting conditions as they only encode appearance tied with specific illuminations. In this work, we propose using object-centric neural scattering functions (OSFs) as object representations in a model-predictive control framework. OSFs model per-object light transport, enabling compositional scene re-rendering under object rearrangement and varying lighting conditions. By combining this approach with inverse parameter estimation and graph-based neural dynamics models, we demonstrate improved model-predictive control performance and generalization in compositional multi-object environments, even in previously unseen scenarios and harsh lighting conditions.

\end{abstract}

\section{Introduction}
\label{sec:intro}

Predictive models are the core components of many robotic systems for solving inverse problems such as planning and control. Physics-based models built on first principles have shown impressive performance in domains such as drone navigation~\cite{bristeau2011navigation} and robot locomotion~\cite{kuindersma2016optimization}.
However, such methods usually rely on complete \textit{a priori} knowledge of the environment, limiting their use in complicated manipulation problems where full-state estimation is complex and often impossible.
Therefore, a growing number of approaches alternatively propose to learn dynamics models directly from raw visual observations~\cite{finn2017visual, ebert2018visual, hafner2019dream, hoque2020visuospatial, babaeizadeh2021fitvid, tian2022benchmark}. 
 
\begin{figure}[t]
  \centering
  \includegraphics[width=1.0\linewidth]{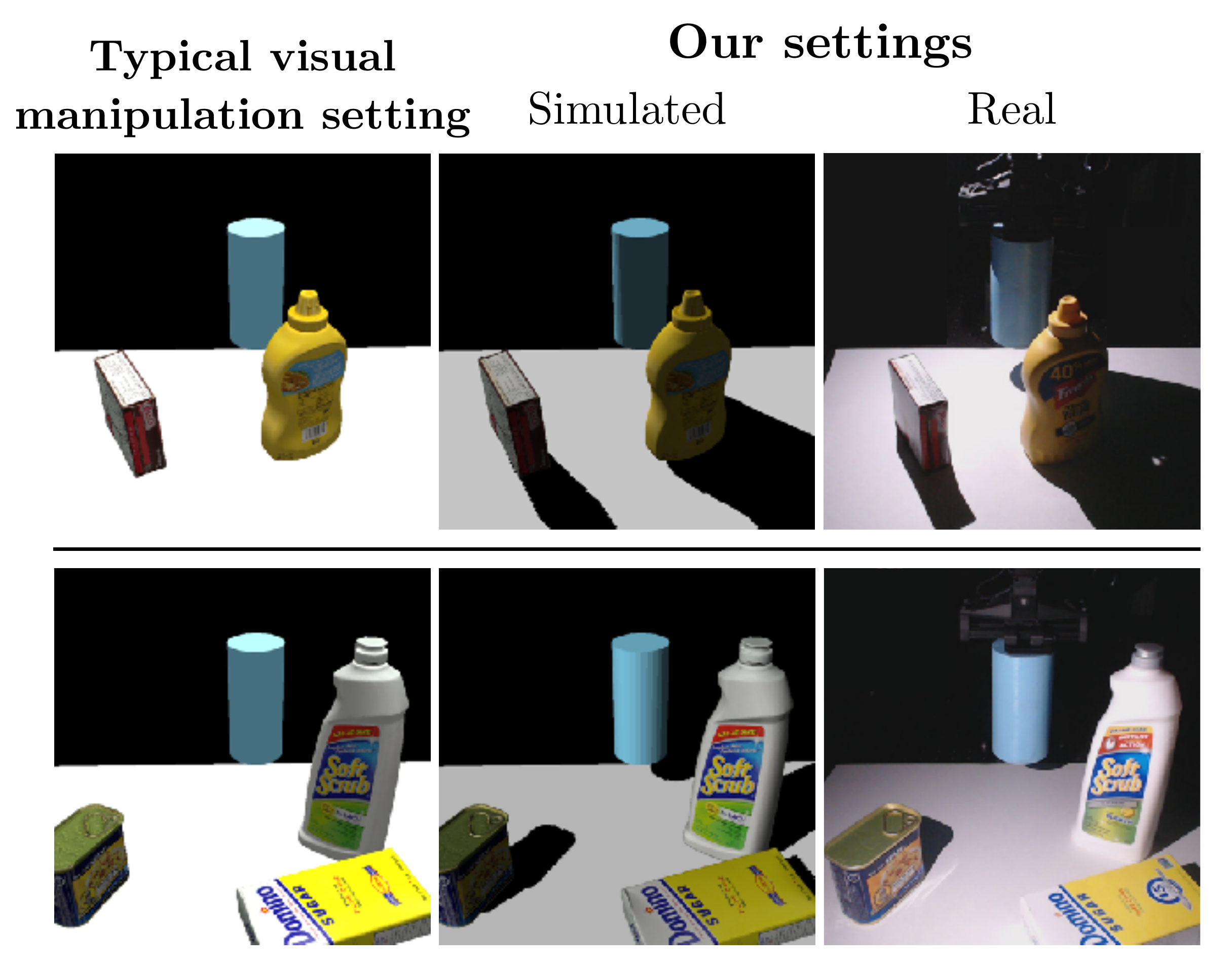}
  \caption{While typically studied visual manipulation settings are carefully controlled environments, we consider scenarios with varying and even harsh lighting, in addition to novel object configurations, that are more similar to real-world scenarios.}
  \label{fig:teaser}
\end{figure}

Although using raw sensor measurements as inputs to predictive models is an attractive paradigm as they are readily available, visual data can be challenging to work with directly due to its high dimensionality. 
Prior methods proposed to learn dynamics models over latent vectors, demonstrating promising results in a range of robotics tasks~\cite{hafner2019learning,schrittwieser2020mastering,hafner2019dream,wu2022daydreamer}.
However, with multi-object interactions, the underlying physical world is 3D and compositional. Encoding everything into a single latent vector fails to consider the relational structure within the environment, limiting its generalization outside the training distribution.

Another promising strategy is to build more structured visual representations of the environment, including the use of particles~\cite{li2018learning,li2020visual,lin2022learning}, keypoints~\cite{manuelli2019kpam,manuelli2020keypoints,li2020causal}, and object meshes~\cite{huang2022mesh}.
Among the structured representations, Driess \etal~\cite{driess2022learning} leveraged compositional neural implicit representations in combination with graph neural networks (GNNs) for the dynamic modeling of multi-object interactions. The inductive bias introduced by GNNs captures the environment's underlying structure, enabling generalization to scenarios containing more objects than during training, and the neural implicit representations allow precise estimation and modeling of object geometry and interactions.
However, Driess \etal~\cite{driess2022learning} only considered objects of uniform color in well-lit scenarios. It is unclear how the method works for objects with more complicated geometries and textures. The lack of explicit modeling of light transport also limits its use in scenarios of varying lighting conditions, especially those vastly different from the training distributions.

In this paper, we propose to combine object-centric neural scattering functions (OSFs)~\cite{yu2023osf} and graph neural networks for the dynamics modeling and manipulation of multi-object scenes. OSFs explicitly model light transport and learn to approximate the cumulative radiance transfer, which allows relighting and inverse estimation of scenes involving multiple objects and the change of lights, such as those shown in Figure~\ref{fig:teaser}.
Combined with gradient-free evolutionary algorithms like covariance matrix adaption (CMA), the learned neural implicit scattering functions support inverse parameter estimation, including object poses and light directions, from visual observations. Based on the estimated scene parameters, a graph-based neural dynamics model considers the interactions between objects and predicts the evolution of the underlying system.
The predictive model can then be used within a model-predictive control (MPC) framework for downstream manipulation tasks.

Experiments demonstrate that our method performs more accurate reconstruction in harsh lighting conditions compared to prior methods, producing higher-fidelity long horizon prediction compared to video prediction models. When combined with inverse parameter estimation, our entire control pipeline improves on simulated object manipulation tasks in settings with varying lighting and previously unseen object configurations, compared to performing MPC directly in image space.

We make three contributions.
First, the use of neural scattering functions supports inverse parameter estimation in scenarios with challenging and previously unseen lighting conditions.
Second, our method models the compositionality of the underlying scene and can make long-term future predictions about the system's evolution to support downstream planning tasks.
Third, we conduct and show successful manipulation of simulated multi-object scenes involving extreme lighting directions.

\section{Related Work}
\label{sec:relat}
\paragraph{Implicit object models for robotic control.}
Neural radiance fields (NeRFs) have emerged as powerful implicit models of scene~\cite{mildenhall2021nerf} and object~\cite{yu2022unsupervised,yang2021objectnerf, jang2021codenerf,smith2022unsupervised} appearance. 
In robotics, NeRFs have been applied to tackle manipulation problems such as grasp and rearrangement planning~\cite{IchnowskiAvigal2021DexNeRF,qureshi2021nerp, jiang2021synergies, danielczuk2021object}, determining constraints and collisions~\cite{ha2021learning, adamkiewicz_vision-only_2022}, learning object descriptors for manipulating object poses~\cite{simeonovdu2021ndf}, and system identification and trajectory optimization~\cite{cleac2022differentiable}. They have also been used as decoders in latent representation learning for model-free~\cite{22-driess-NeRF-RL-preprint} and model-based~\cite{li20213d} reinforcement learning. 
While these prior methods have explored the use of NeRFs for robotic manipulation, they train implicit models that are either restricted to baked-in lighting settings determined at training time or model the entire scene together; therefore, these methods cannot handle compositionality. In this work, we aim to leverage object-centric, \textit{relightable} neural scattering functions in a model-based planning framework. 

\paragraph{Pose estimation from images.}
Many prior works investigate the problem of estimating rigid object poses from RGB images, including deep-learning approaches based on correspondences ~\cite{tekin18, zakharov2019dpod,jafari2018ipose,li2019cdpn,park2019pix2pose,peng2019pvnet} as well as direct regression~\cite{xiang2018posecnn,zhou2019objects,engelmann2021points,labbe2020cosypose,deng2021poserbpf}. 
While achieving impressive results, these methods require colored 3D object meshes to train on and are sensitive to changing lighting conditions. Yen-Chen \etal~\cite{yen2020inerf} and Jang \etal~\cite{jang2021codenerf} study the problem of estimating the camera pose of a given image using neural fields. In this work, we propose estimating both object poses \textbf{and} lighting conditions from RGB images by ``inverting'' implicit OSF models. 

\paragraph{Visual planning with learned dynamics models.}
Learned dynamics models are a powerful tool for performing planning. When the agent receives observations in 2D images, one class of prior works directly models dynamics in image space~\cite{finn2017visual}. Another approach is to learn a keypoint~\cite{manuelli2020keypoints} or latent state representation via reconstruction by a convolutional neural network~\cite{ha2018world, hafner2020mastering} or volumetric rendering~\cite{li20213d}. Alternatively, scenes can be represented explicitly as discrete objects with learned decoding or rendering modules~\cite{janner2019reasoning, veerapaneni2020entity}. While our approach also explicitly models a scene as a set of discrete objects, we perform inverse pose estimation to acquire the 6D pose of each object in the scene. Combined with learned OSFs, we can compose objects to render the scene.

\label{sec:method}
\section{Problem Definition}
Given a set of $N$ known objects with pre-trained OSFs models, a goal image depicting the desired configuration of the objects $I_\text{goal}$, and RGB camera observations of the scene from $V$ different viewpoints at each timestep $t$, denoted $I_t^{1:V}$, the objective is to execute a sequence of actions $a_{0:T}$ such that the goal object configuration is achieved at the end of $T$ steps. Each action in the action sequence is defined as a $3$D change in position for a cylindrical pusher object that is present in all scenes, but could be generalized to represent many types of robot end-effectors. We assume access to the camera parameters for each observation and the goal image, but \textit{not} the lighting parameters.

While we assume access to pre-trained OSFs models for each object, we note that multi-view images of each object with paired light pose information are sufficient supervision for training OSFs. We do not require access to object meshes. In this work, we constrain ourselves to handle only rigid objects for simplicity.

\section{Methods}
To solve tabletop visual object manipulation tasks, we want our model to have two properties:
\begin{enumerate}[itemsep=2pt, topsep=5pt, partopsep=0pt, parsep=2pt]
    \item It should have a compact, low-dimensional representation of the scene, which allows learned dynamics models to perform more stable long-horizon prediction.
    \item It should be capable of handling scenarios with different object configurations and previously unseen, unknown lighting.
\end{enumerate}

We achieve the first property by representing scenes using the $6$D pose of each object. Then, we achieve the second by implicitly modeling each object using object-centric neural scattering functions (OSFs). OSFs model the light transport for a particular object -- specifically, given the spatial location and incoming and outgoing light direction, they predict the object's radiance transfer from the light source to the viewer. This enables relighting as well as composition during the rendering process. 

Our method consists of three main steps. The first is an offline optimization phase, where we train a dynamics model using simulated data that takes as input $6$D object poses of each object in the scene and the pusher's action, and outputs future $6$D poses for each object. Then, at inference time, we use inverse parameter estimation with OSFs models to estimate the scene's initial object poses and the light position. Finally, we use the learned dynamics model to perform model-predictive control, updating object states at each step using the same inverse parameter estimation strategy.
We introduce each component of our method in the following sections and present a summary in Algorithm~\ref{alg1}.

\begin{figure}[t]
  \centering
  \includegraphics[width=1\linewidth]{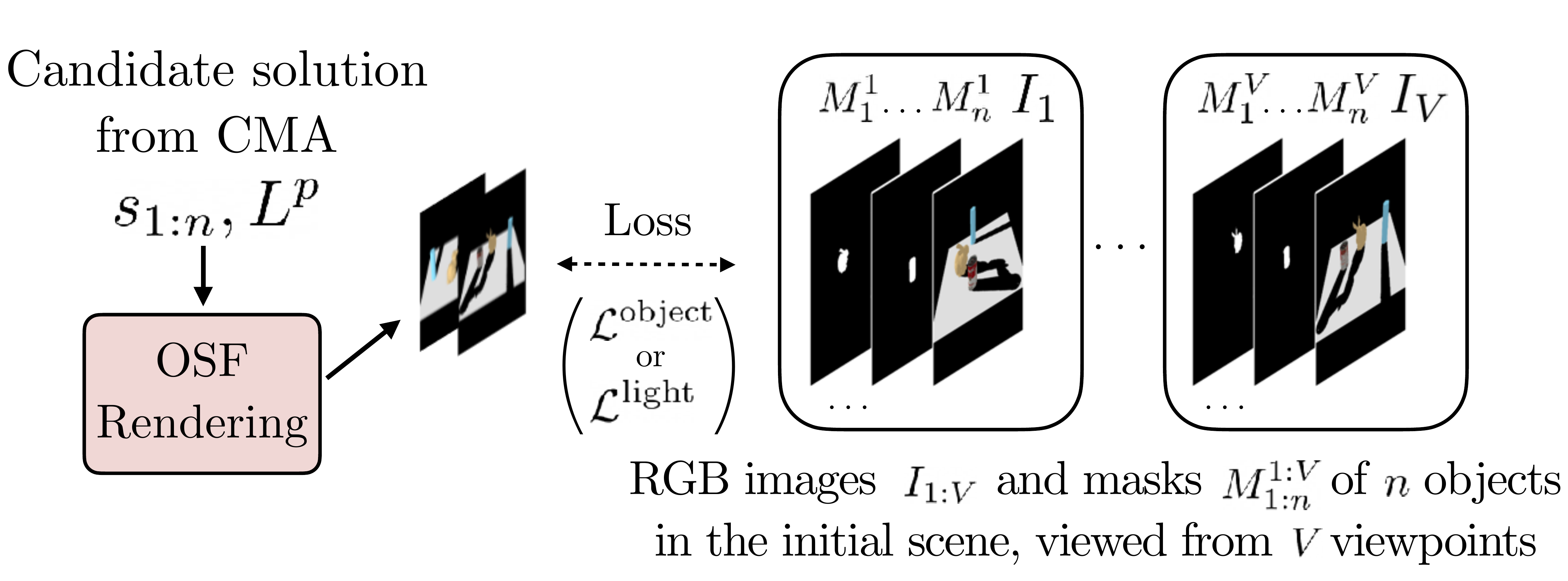}

  \caption{Pose estimation for the light and objects. Given multi-view images of the initial static scene and the binary masks of each object, we use CMA~\cite{6790790} to estimate the initial 6D pose (position and quaternion) of each object and the 3D position of the light source. Unlike previous methods, we explicitly model the lighting to better recover poses in novel lighting directions.}
  \label{fig:pose estimation1}
\end{figure}

\subsection{Neural Implicit Scattering Functions}
\label{sec:neural_implicit_scattering_func}
Since we use object poses to represent the scene, we need to estimate them from visual observations. To this end, we \textit{invert} the rendering process of the scene. Since we would like our estimation approach to generalize to any configuration of objects, we adopt an object-centric, composable implicit model. Specifically,
we use the object-centric neural scattering functions (OSFs) \cite{yu2023osf} to represent each object. 

OSFs are relightable, compositional implicit neural rendering models based on neural radiance fields (NeRFs)  \cite{mildenhall2021nerf}. To achieve relighting, OSFs learn to approximate the cumulative radiance transfer from a distant light in addition to learning the spatial volume density as NeRFs do:
\begin{equation}
f_\theta: (\mathbf{x}, \boldsymbol{\omega}_\text{light}, \boldsymbol{\omega}_\text{out})\rightarrow({\rho}, \sigma),
  \label{eq:OSF1}
\end{equation} 
where $\mathbf{x}$ denotes the 3D spatial location, $\sigma$ denotes the spatial volumetric density, $\boldsymbol\omega_{\text{out}}$ denotes the outgoing radiation direction, $\boldsymbol\omega_{\text{light}}$ denotes the distant light direction and $f_\theta$ denotes a learnable deep neural network. $\rho(\mathbf{x}, \boldsymbol\omega_{\text{light}}, \boldsymbol\omega_{\text{out}})$ denotes the cumulative radiance transfer function. The scattered outgoing radiance $L_{\text{out}}$ can then be formulated as
\begin{equation}
L_\text{out}(\mathbf{x}, \boldsymbol{\omega}_\text{out}) = \int_{S^2} \rho(\mathbf{x}, \boldsymbol{\omega}_\text{light}, \boldsymbol{\omega}_\text{out}) L_\text{light}(\boldsymbol{\omega}_\text{light})d\boldsymbol{\omega}_\text{light},
  \label{eq:OSF2}
\end{equation} 
where $S^2$ denotes the unit sphere for integrating solid angles, and $L_{\text{light}}$ denotes the radiance of the distant light. 

To accelerate the rendering of OSFs, we use a variant called KiloOSFs~\cite{yu2023osf}. KiloOSFs extend the idea of KiloNeRFs~\cite{reiser2021kilonerf}, which represent a static scene as thousands of small independent MLPs, to the object-centric setting.
For each pixel depicted by a ray $\mathbf{r}(t)=\mathbf{o}-t\boldsymbol{\omega}_\text{out}$, we use compositional volumetric rendering for the scene:
\begin{equation}
L(\mathbf{o}, \boldsymbol{\omega}_\text{out}) = \int_{t_n}^{t_f} T(t)\sigma^\text{s}(\mathbf{r}(t))L_\text{out}^\text{s}(\mathbf{r}(t),\boldsymbol{\omega}_\text{out})dt,
  \label{eq:OSF3}
\end{equation}
where $T(t)$ denotes the accumulated transmittance along the ray from the near plane $t_{n}$ to the far plane $t_{f}$, $\sigma^\text{s}$ denotes the sum of density from all objects in the scene, and $L_\text{out}^\text{s}$ denotes the sum of radiance from all objects. For inference speed, we render shadows with shadow mapping~\cite{williams1978casting}.

\subsection{Inverse Parameter Estimation}
\label{sec:inverse_param_est}

\begin{figure}[t]
  \centering
  \includegraphics[width=1\linewidth]{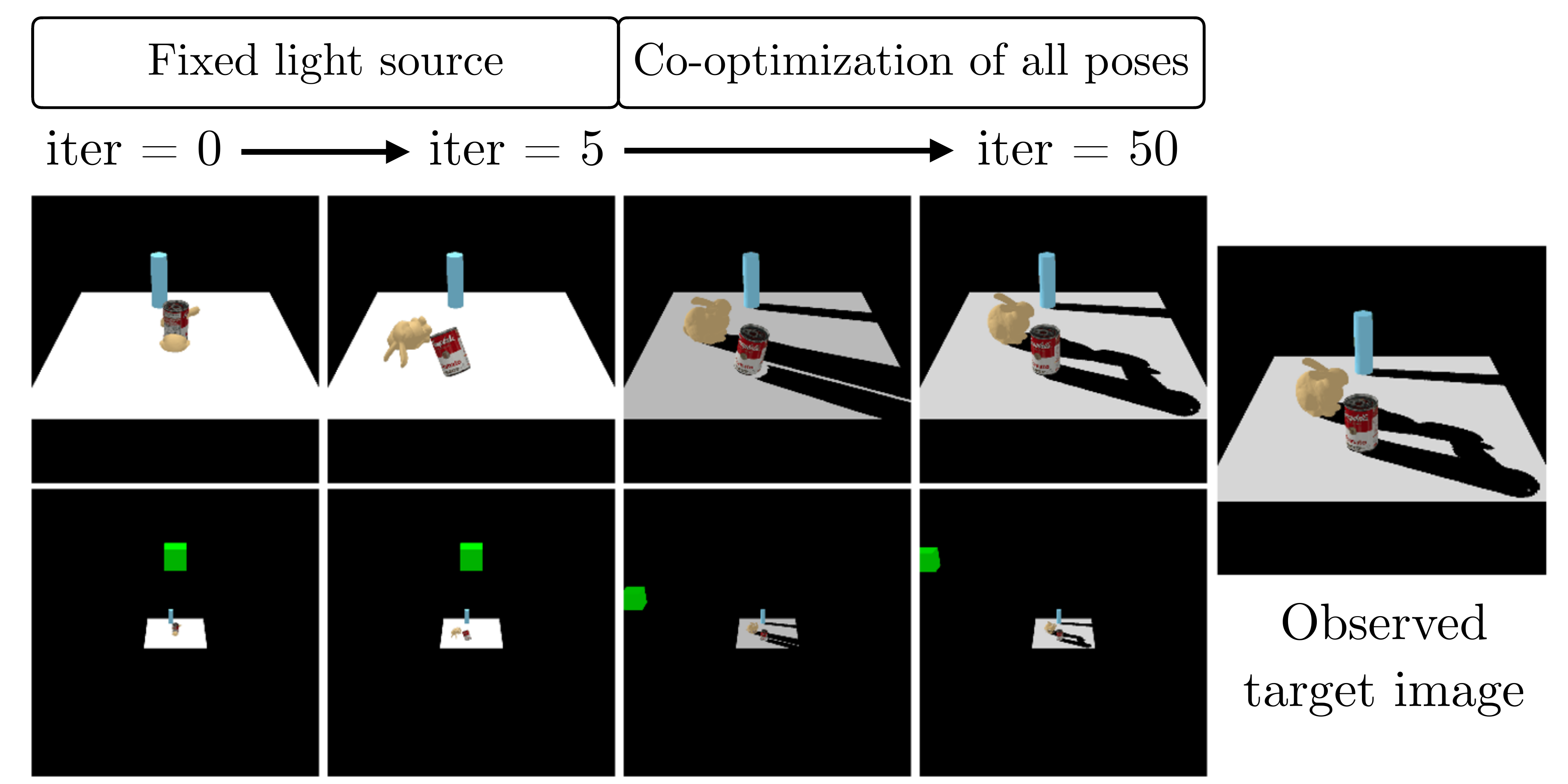}
  \caption{Demonstration of inverse parameter estimation. The top and bottom rows show the \textbf{same} tabletop setting from close and far views, respectively. The \textcolor{ForestGreen}{green cube} represents the light position, which always points at the plane's center. In this figure, for the first five iterations, we fix the light pose directly overhead and optimize all object poses. When object poses are approximately optimized, we estimate all object poses and the light pose together.
  }
  \label{fig:pose estimation2}
\end{figure}

To increase the stability of long-term dynamics predictions, we use coordinate space pose vectors (rather than latent vectors) to represent objects compactly.  Following Driess et al.~\cite{driess2022learning}, we assume access to RGB images of the initial scene $I_i \in \mathbb{R} ^ {H \times W \times 3}$, where $i = 1,...,V$ represents the index into $V$ camera views, as well as binary masks $M_{j}^{i} \in\{0,1\}^{H \times W}$ of each object $j$ in view $i$. We also assume that we already have KiloOSF models $O_j, j=1,...,n$ for each object in the scene. We refer to the KiloOSF models for the compositional rendering of \textit{multiple} objects (including shadows) as $O^{all}_{1:n}$. For the rest of this section, we use the notation $O_j(v, s, L)$ to indicate rendering object $j$ with pose $s$ from viewpoint $v$ with directional light pose $L$.

We use covariance matrix adaptation (CMA) \cite{6790790}, represented by  $\Psi$, to optimize the 6D poses of each of $n$ objects $s_{1:n} \in \mathbb{R} ^ {n \times 7}$ and light position $L^p \in \mathbb{R} ^ {3}$:

\begin{equation}
s_{1:n}, L^p=\Psi\left(I_{1: V}, M_{1: n}^{1: V}, O_{1:n}\right),
  \label{eq:pose1}
\end{equation}
as shown in Figure \ref{fig:pose estimation1}. For readability, we write Equation~\ref{eq:pose1} as a function of OSFs $O_{1:n}^{all}$; in practice, the OSFs are used to define the loss functions as described below.

For object pose optimization when given object masks, we use the mean-squared error (MSE) of $V$ multi-view KiloOSF-rendered images of the object and the observed multi-view images $I_{1:V}$, masked by binary masks for each object $M^{1:V}_j$. We initialize the light direction $L_{init}$ as the unit vector in the $z$-direction. Specifically,
\begin{equation}
\mathcal{L}  ^\text{object}_j = \sum_{v\in V} ||O_j(v, s_j, L_{init}) - (I_{v} \circ  M^{v}_j)||^2_2.
  \label{eq:pose2}
\end{equation} 
For light pose optimization, we compositionally render the entire scene, including shadows, with KiloOSF $O^{all}_{1:n}$
and compute the MSE with the observed images:
\begin{equation}
\mathcal{L} ^\text{light} = \sum_{v\in V}||(O^{all}_{1:n}(v, s_{1:n}, L^p),  - I_{v}||^2_2.
  \label{eq:pose2}
\end{equation} 
Figure \ref{fig:pose estimation2} shows a visualization of the process.

\subsection{Action-Conditioned Dynamics Model}
\label{sec:dynam}

During dynamics prediction, we are given object poses and an action taken by an agent in the scene, and aim to infer subsequent states. Concretely, we train a graph neural network (GNN) dynamics model to make predictions of the future object $6$D poses:
\begin{equation}
s_{1:n}^{t+1} = f_{GNN}(s_{1:n}^{t}, A^t, a^t) \in \mathbb{R}^{n\times7}, 
  \label{eq:GNN1}
\end{equation} 
where $A^t \in \{0,1\}^{n,n}$ is the adjacency matrix, $a^t \in \mathbb{R} ^3$ is the input action, and $s_{1:n}^{t} \in \mathbb{R} ^{n \times 7}$ are poses of all objects at time $t$. To model multi-object interactions, we follow 
 Li \etal ~\cite{li2019propagation} and perform multiple inter-object propagation steps during the prediction for a single future time step.

We use each node in a graph to represent a single object, with its pose $s^t$ as its input node feature. Following \cite{driess2022learning}, we dynamically create graph edges to improve the stability of long-term predictions. We only construct edges between nodes if the objects represented by those nodes can potentially collide during the current timestep. We represent edge information using an adjacency matrix $A^t$. Because we are working in coordinate space, we use an approximated distance between two objects to determine potential collisions by leveraging the geometric information present in KiloOSF models. We first construct approximate point clouds by evaluating the KiloOSF's density on a grid of points and thresholding points above a certain density. Then, we use the longest axis of the tightest bounding box containing the entire pointcloud as the threshold distance $\kappa$ for edge creation. Note that the creation of an edge does not mean that the model \textit{must} predict a collision, but rather that it \textit{can}.
\begin{equation}
A^t_{i j}=\left\{\begin{array}{ll}
1 & ||s_{i}^{t} - s_{j}^{t}||^2_2 < \kappa, i \ne j \\
0 & \text { else }
\end{array}\right. 
  \label{eq:GNN2}
\end{equation}

We use a shared edge encoder $\mathbb{E}^\text{edge}$ and propagator $\mathbb{P}^\text{edge}$ regardless of which objects are present in the scene. This greatly enhances the generalization ability of GNN to handle unseen scenarios while using more training data. The detailed structure of the GNN is shown in the Appendix.

We train the GNN dynamics model using a mean squared-error loss between the predicted and ground truth object poses, each represented by a position $p \in \mathbb{R}^3$ and unit quaternion $q \in \mathbb{R}^4$, summed across all objects in the scene. Dynamics models trained on single-step prediction are prone to compounding errors during longer open-loop model rollouts. Thus, we train the model on predictions of up to $3$ future timesteps. 

\subsection{Visual Model-Predictive Control}
\label{sec:mpc}
\begin{figure*}[t]
  \centering
  \vspace{-1em}
  \includegraphics[width=0.95\linewidth]{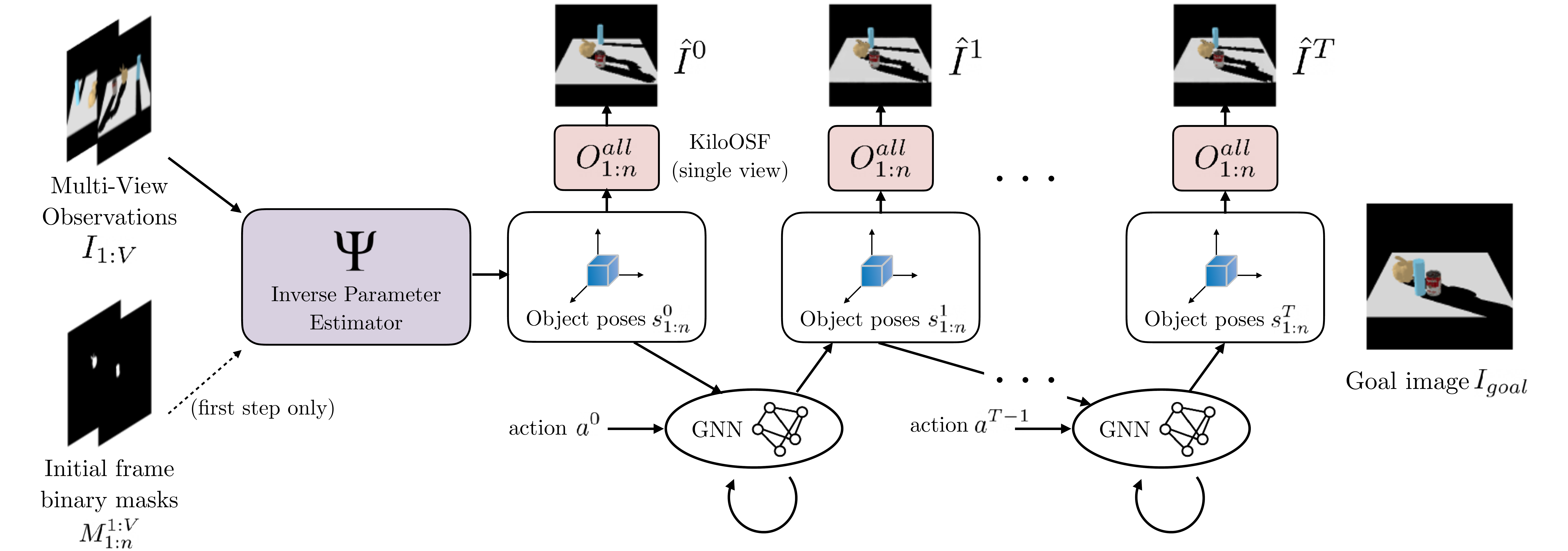}

  \caption{Combined framework for perception (inverse parameter estimation) and prediction (dynamics model). The figure shows the $r$-th replanning step in model-predictive control. In the first step, we use the multi-view images $I_{1:V}$ and the object's binary masks $M^{1:V}_{1:n}$ for inverse parameter estimation. Afterward, only multi-view images are used. We only render the predicted images $\hat{I}^{t:t+H}$ using KiloOSF with the \textit{goal image view}, and then compare the sequence of images $\hat{I}^{t:t+H}$ with the goal image $I_{goal}$ via an MSE loss.}
  \label{fig:frame1}
\end{figure*}

Given a goal image and initial visual observations of an environment, the objective of visual model-predictive control is to optimize a robot action sequence to reach the goal. 

After estimating the light position and each object's initial pose, we perform sampling-based planning using our learned dynamics model. At timestep $t$, we first sample ${K}$ action sequences of length $H$. By rolling out open-loop predictions from the GNN dynamics model, we obtain the predicted future poses of each object $\hat{s}^{t:t+H}_{1:n}$ for each sample.
We then provide these objects and light poses to KiloOSF to render RGB images $\hat{I}^{t:t+H}$. We use the squared $\ell_2$ error between rendered images and the goal image, that is, $\sum_{\tau=t}^{t+H} ||\hat{I}^{\tau} - I_{goal}||^2_2$ as the cost function to score each action sequence sample as in visual foresight\cite{ebert2018visual}. We then use MPPI~\cite{williams2015model} to update the action sampling distribution based on the action samples and scores. The first step of the action distribution's mean is executed in the environment. 

Then, the object pose estimates $s^{t+1}_{1:n}$ are updated again using inverse parameter estimation using new observations from the environment $I_{1:V}$. To reduce computational cost, we decrease the search space for inverse parameter estimation at this step by reducing the initial standard deviation of CMA compared to the initial step. We also alleviate the assumption of object masks past the initial frame, and compute the loss for the CMA optimizer as the mean-squared error (MSE) between rendered and observed RGB frames.

The process is repeated to replan every $r$ time steps. Algorithm \ref{alg1} outlines the entire planning procedure for $r=1$, and Figure~\ref{fig:frame1} visualizes a single (re-)planning step.

\begin{algorithm}
	\renewcommand{\algorithmicrequire}{\textbf{Input:}}
	\renewcommand{\algorithmicensure}{\textbf{Output:}}
	\caption{Visual MPC with OSFs and GNN dynamics}
	\label{alg1}
	\begin{algorithmic}[1]
        \REQUIRE Pre-trained KiloOSF models $O_{1:n} (O^{all}_{1:n})$, initial environment image observation ${ }I_{1:V}$, initial object masks $ M_{1: n}^{1: V}$, goal image $I_{goal}$ and viewpoint $v_{goal}$, GNN dynamics model $f_{GNN}$, planning horizon $H$
        \FOR{ $t = 0...T$ }
            \IF{ $t = 0$} 
                \STATE // Perform initial object \& light pose estimation
                \STATE ${ }s^0_{1:n}, L^p=\Psi\left({ }I_{1: V}, M_{1: n}^{1: V}, O_{1:n}\right)$
            \ELSE
                \STATE // Refine object pose estimates without masks
                \STATE $s^t_{1:n} = \Psi\left(I_{1: V}, O^{all}_{1:n}\right)$
            \ENDIF
                \STATE Sample $K$ random action sequences ${a_{{1:K}}^{t:t+H}}$.
                \FOR { action sample $a^{t:t+H}$ in $a_{{1:K}}^{t:t+H}$ }
                    \STATE // Iteratively predict $H$ future states with the action-conditioned GNN dynamics model.
                    \STATE $\hat{s}^{t:t+H}_{1:n} = f_{GNN} (s^{t}_{1:n}, a^{t:t+H})$.
                    \STATE // Use KiloOSF to render image predictions from the goal image viewpoint. 
                    \STATE $\hat{I}^{t:t+H} = O^{all}_{1:n}(v_{goal}, \hat{s}^{t:t+H}_{1:n}, L^p)$ .
                    \STATE // Compute loss for each sampled action sequence.\STATE $L = \sum_{\tau=t}^{t+H} ||\hat{I}^{\tau} - I_{goal}||^2_2$.
                \ENDFOR
                \STATE Update action sampling distribution via MPPI  \cite{williams2015model}.
                \STATE Execute the first step of the mean of the updated action sampling distribution.
                \STATE Receive a new environment observation $I_{1:V}$.
        \ENDFOR
	\end{algorithmic}  
 \vspace{-0.05cm}
\end{algorithm}
\vspace{-0.5em}

\section{Experiments}
\label{sec:exper}
\begin{figure*}[t]
  \centering
  \vspace{-1em}
  \includegraphics[width=1\linewidth]{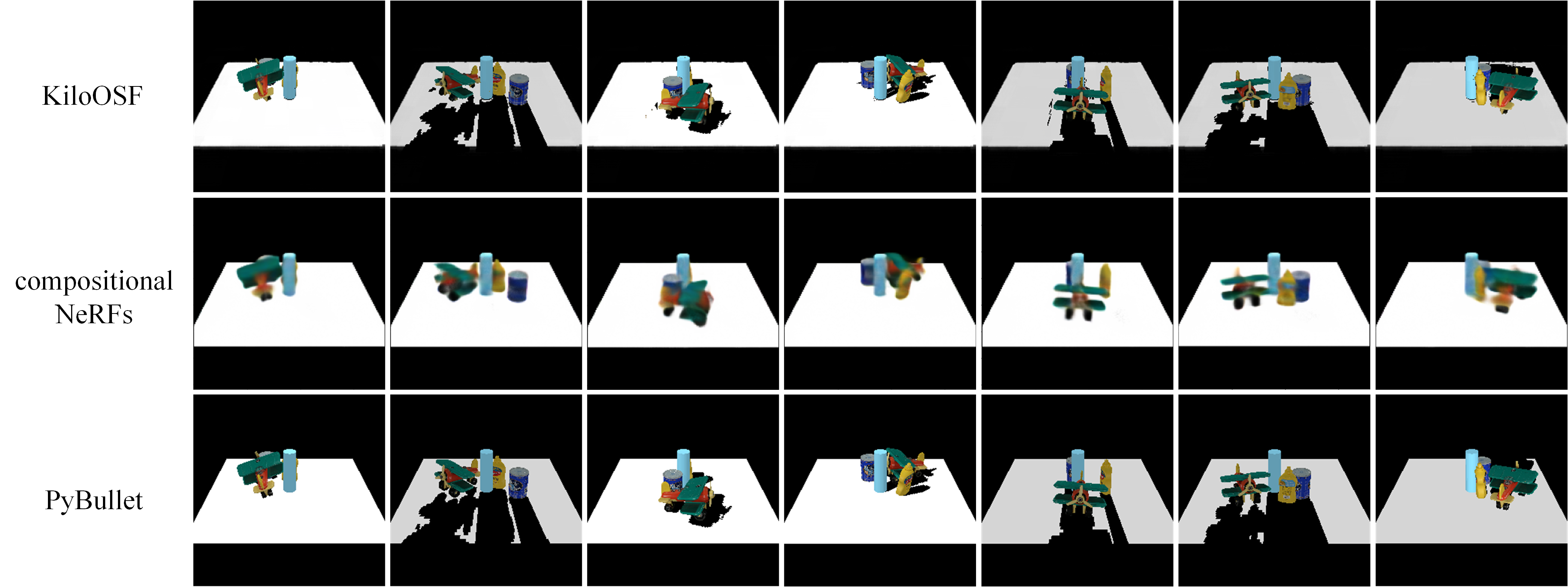}

  \caption{Qualitative comparison of the reconstruction results of KiloOSF compared to compositional NeRFs~\cite{driess2022learning}. The first row shows images rendered by our KiloOSF, and the second row shows the visual reconstruction of compositional NeRFs. The third row shows the ground truth image rendered directly with PyBullet. Our method can reasonably render the color change of the object due to the change of lighting pose, such as the plane becoming darker when the light is tilted. At the same time, our method can render accurate shadows.}
  \label{fig:reconstruction}
\end{figure*}

In this section, we empirically evaluate the performance of our method and its components. Specifically, we seek to answer the following questions:
\begin{enumerate}[itemsep=2pt, topsep=5pt, partopsep=0pt, parsep=2pt]
    \item How well can KiloOSF reconstruct scenes with composable objects in harsh lighting conditions?
    \item Does the combination of KiloOSF and the GNN dynamics model enable long-horizon visual dynamics prediction?
    \item How does our proposed method compare to existing methods for performing robotic manipulation using visual model-predictive control?
    \item How well can our method generalize to performing control in settings unseen at training time?
    \item Can our lighting and pose estimation methods be applied to real-world settings? 
\end{enumerate}

\subsection{Experimental Setting}
We focus on settings with particularly harsh lighting conditions to determine whether our method can still perform manipulation when lighting causes significant appearance changes in the scene.
We use the PyBullet simulator~\cite{coumans2016PyBullet} to perform experiments. We consider a tabletop manipulation setting, shown in Figure~\ref{fig:pose estimation2}. This consists of a cylindrical pusher, representing the end-effector of a robot, one or more objects to be manipulated, and four camera viewpoints. Compared to previous methods, our object representations can handle objects with finer textures. Therefore, we use a subset of $20$ YCB objects \cite{calli2015ycb, calli2015benchmarking} in our experiments.

To test the generalization ability of our model, we train our dynamics model in scenes with only three objects, along with the pusher. However, during testing, in addition to the three-object setting, we also test on scenes with two and four objects, as described in Section \ref{sec 5.5}.

\subsection{Visual Reconstruction}
Multi-object dynamic interaction models, such as graph neural networks, are effective for making predictions in coordinate or latent space rather than in image space. However, inferring these low-dimensional states from images, for example through inverse parameter estimation, requires faithful image reconstruction from low-dimensional states.

\begin{figure*}[t]
  \centering
  \vspace{-1em}
  \includegraphics[width=1\linewidth]{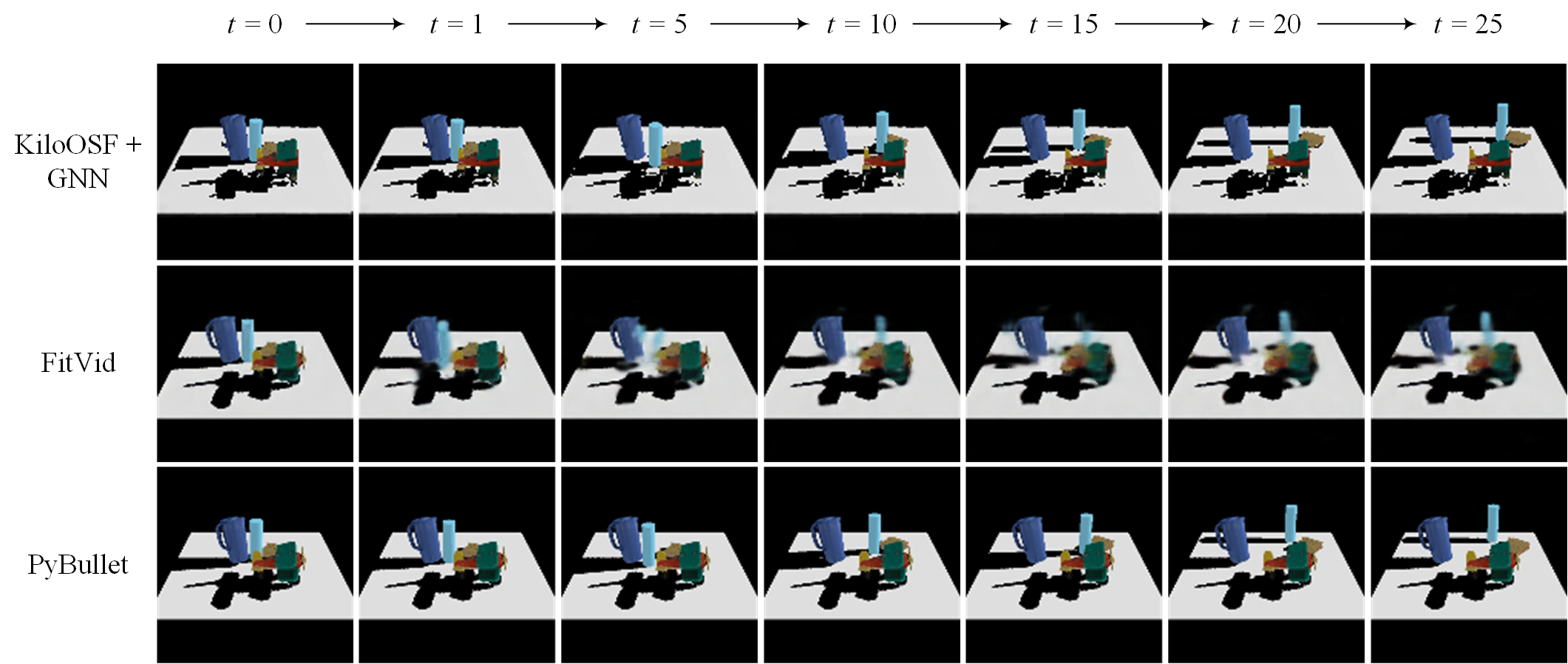}

  \caption{Qualitative comparison of the forward prediction with our method compared to FitVid, a video prediction model. 
  Our method achieves much better prediction results than FitVid, even though FitVid is trained on a dataset that already contains multiple lighting poses and shadows. The third row shows the ground truth images for the trajectory from the PyBullet simulator.}
  \label{fig:prediction}
\end{figure*}

Prior work~\cite{driess2022learning} %
uses compositional neural radiance fields (compositional NeRFs) to support the rendering of multi-object composite scenes. For this experiment, we compare KiloOSF reconstructions to those of compositional NeRFs.

We collect $1000$ trajectories of simulated object data to train the compositional NeRF models and our dynamics model. In each trajectory, three YCB objects are randomly selected to be placed into the scene. The cylinder is then moved randomly for $50$ steps with actions drawn from a uniform distribution $[\mathcal{U}(-0.3, 0.3), \mathcal{U}(-0.3, 0.3), 0]$, representing the speed ( $m/s$ ) at which the cylinder moves. Each action is applied for roughly 0.2 seconds. 

In Figure~\ref{fig:reconstruction}, we show qualitative comparisons between the reconstructions generated by our KiloOSF and by compositional NeRFs. 
Although compositional NeRFs can render multi-object scenes, they can suffer, for example, when the scene's lighting changes significantly from that experienced during training. Additionally, compositional NeRFs cannot fit high-frequency information such as fine textures and complex geometry as well as our KiloOSF model. 

\subsection{Visual Prediction}
Next, we evaluate the visual prediction performance of our combined GNN dynamics and KiloOSF rendering modules. We compare to a state-of-the-art stochastic variational video prediction model, FitVid~\cite{babaeizadeh2021fitvid}, that takes RGB images as input and makes predictions directly in pixel space. For fairness, we modify the FitVid model to make predictions at $128 \times 128$ resolution instead of the default $64 \times 64$. We then train FitVid on the same dataset of $1000$ trajectories that we use to train our dynamics model, randomizing the lighting present in each trajectory to improve its performance. 

We visualize qualitative prediction results in Figure~\ref{fig:prediction}. We see that over longer prediction horizons past around $10$ predicted steps, the prediction quality of FitVid quickly deteriorates. This is because small errors in the pixel predictions of the model accumulate rapidly. By separating the dynamics and rendering model, our GNN dynamics model can make long-horizon predictions in low-dimensional coordinate space to be later rendered using KiloOSF, without sacrificing appearance quality.

\begin{figure*}[t]
  \centering
  \vspace{-1.5em}
   \begin{subfigure}{0.33\linewidth}
   \includegraphics[width=\textwidth]{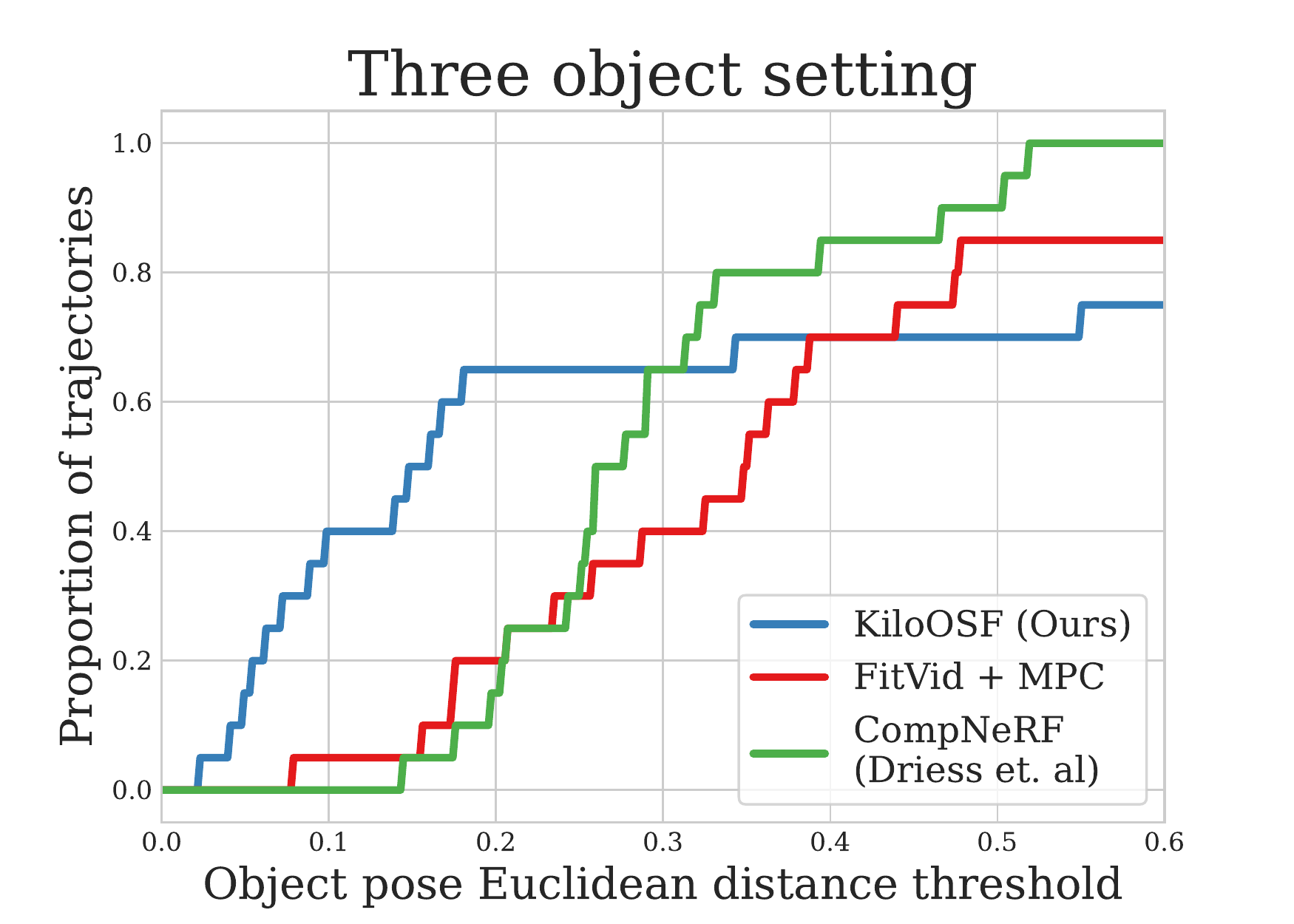}
   \subcaption{Three object (training) scenario.}
   \end{subfigure}
   \begin{subfigure}{0.66\linewidth}
   \includegraphics[width=0.5\textwidth]{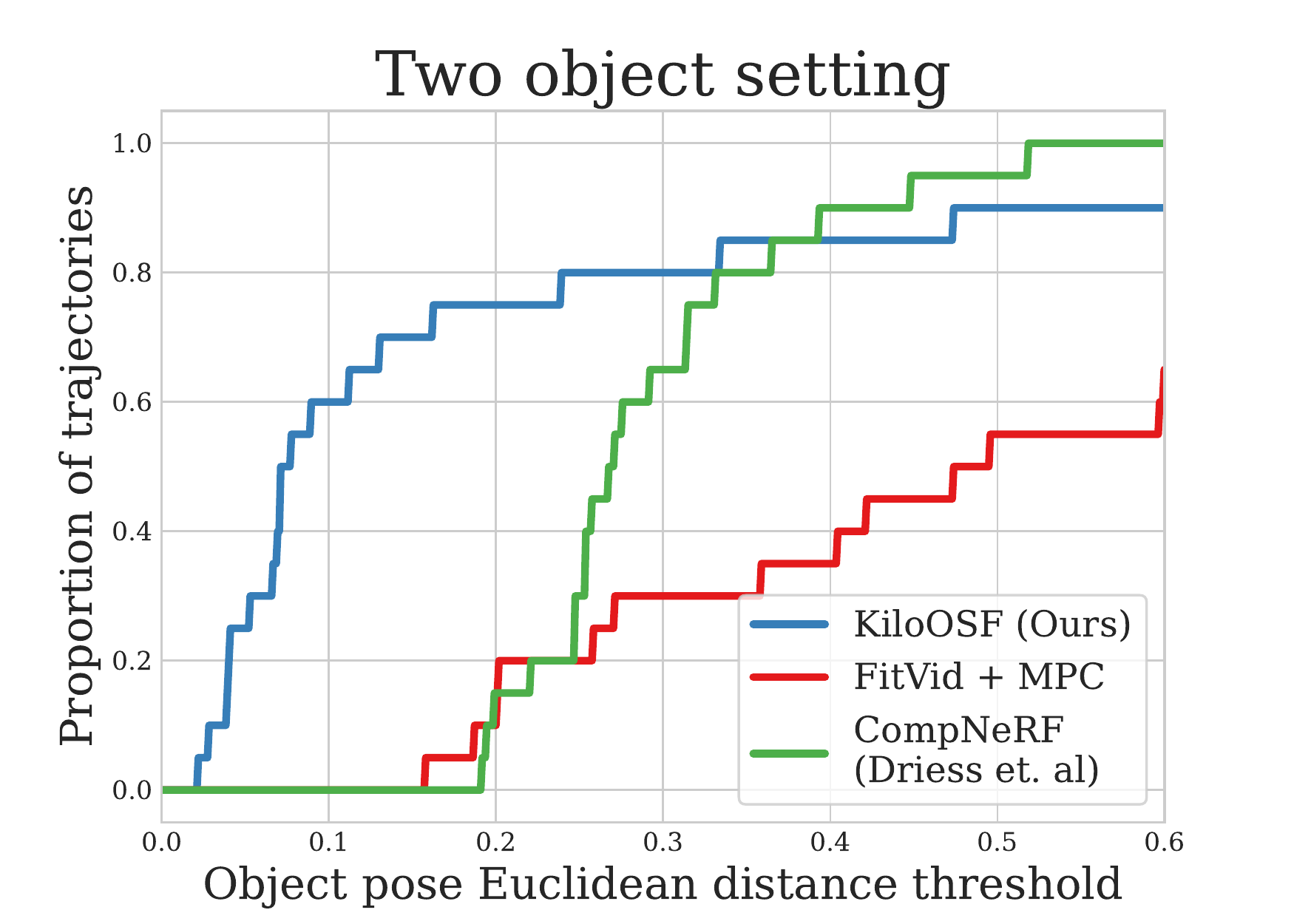}
   \includegraphics[width=0.5\textwidth]{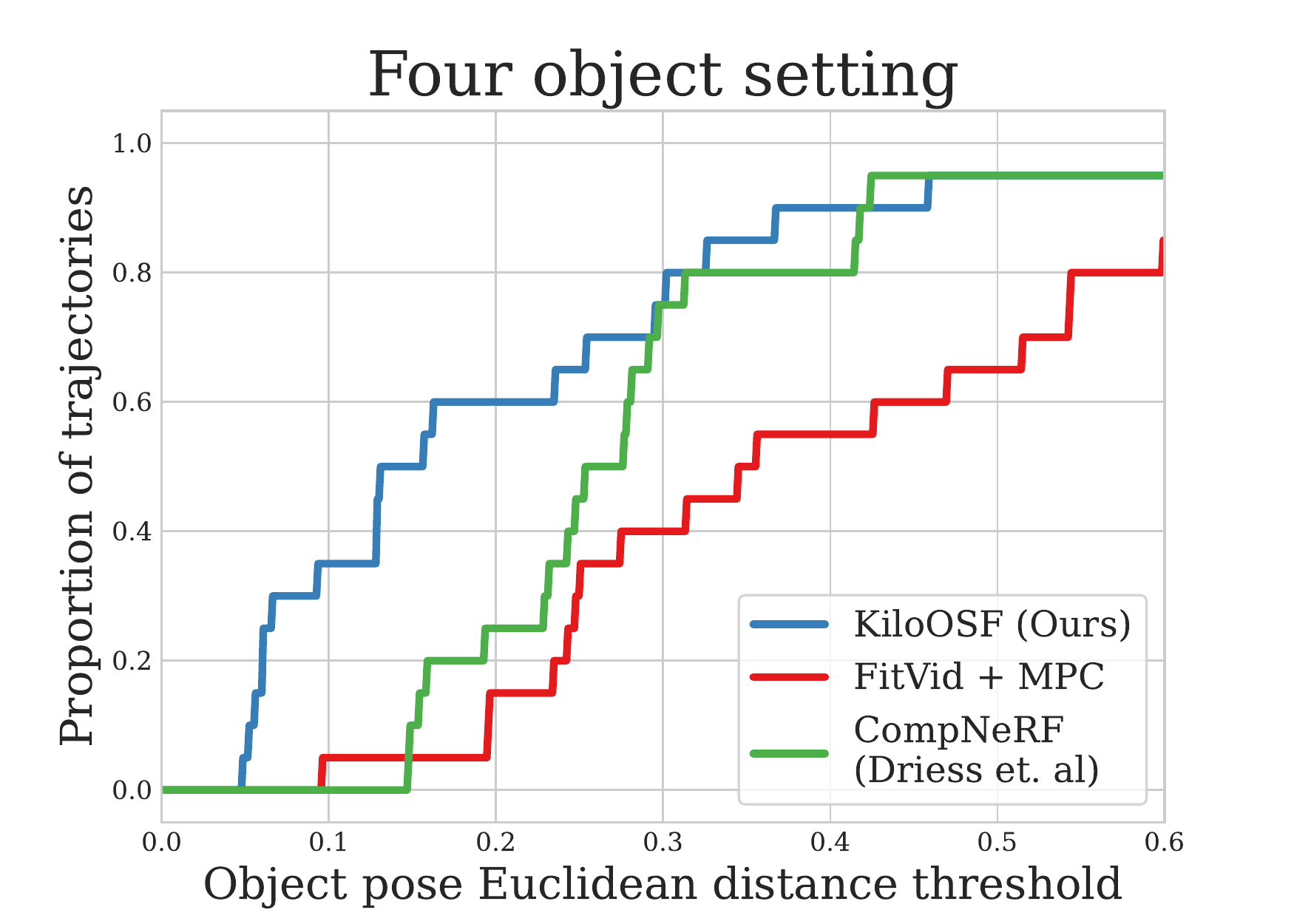}
   \subcaption{Two and four object (unseen) scenarios.}
 \end{subfigure}
  \caption{Quantitative results of our method compared to running visual MPC with FitVid~\cite{babaeizadeh2021fitvid} and Compositional NeRF~\cite{driess2022learning} in experimental settings with two, three, and four objects respectively. The horizontal axis represents a range of loss thresholds. The vertical axis represents the proportion of experimental trials where models rearrange the objects within the given error threshold. After 30 steps of MPC, we compute the Euclidean distance between the 6D object poses at the final step. 
  Our method outperforms the pixel-space MPC and Compositional NeRF-based methods in all three experimental settings. 
  }
  \label{fig:mpc_results}
\end{figure*}

\subsection{Model Predictive Control}
In this section, we evaluate the performance of our method as a full model-predictive control pipeline. This includes performing inverse object pose and lighting estimation, planning and scoring action sequences, and updating estimated parameters in a closed-loop fashion. 

We evaluate our method as well as visual foresight~\cite{finn2017visual} with FitVid as the visual dynamics model on $20$ different testing tasks, each with randomized combinations of three objects and a randomized, unknown light pose sampled in a hemisphere above the ground plane. Each goal image specifies moving one or more objects at a distance of at least $0.075$m from their initial location. This is a challenging setting because the randomized lighting conditions can create harsh lighting conditions, such as those shown in Figure~\ref{fig:prediction}. 

We execute model-predictive control for $30$ steps for each method, and report the performance in terms of the final 6D object pose error (Euclidean distance from ground truth goal pose) over all objects. The results are presented in Figure~\ref{fig:mpc_results}. By inferring the lighting parameters in each trial, our model can better predict the underlying scene dynamics, improving planning performance.

\subsection{Generalization to Unseen Scenarios}
\label{sec 5.5}
Furthermore, we evaluate how our method generalizes to previously unseen scenarios. We consider testing scenes that contain two or four objects, while during training, there are always three objects present. In Figure~\ref{fig:mpc_results}, we present the control performance of the same model across these settings. Because our method factorizes the scene into individual object representations and uses a graph neural network dynamics model for object interactions, it naturally handles this case. For models trained on visual observations of the entire scene, these settings are out of the training distribution. 
For computational reasons we perform evaluation providing object masks at all steps as in Driess et al.\cite{driess2022learning}, but we find that our method achieves similar performance in the three-object setting with and without full masks.

\subsection{Real World Lighting and Pose Estimation}
\begin{figure}[h]
  \centering
  \vspace{-1em}
  \includegraphics[width=1.0\linewidth]{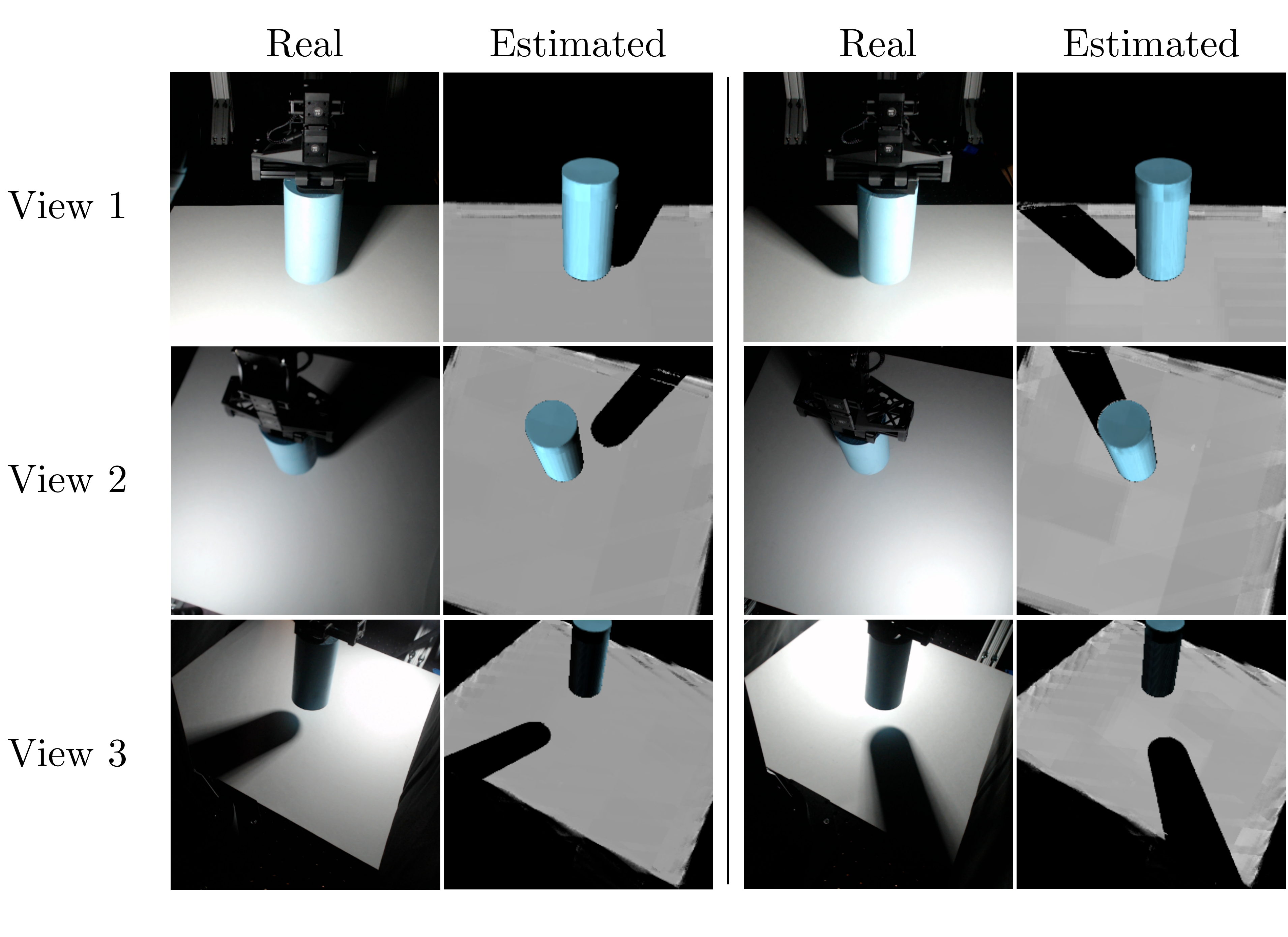}
  \includegraphics[width=1.0\linewidth]{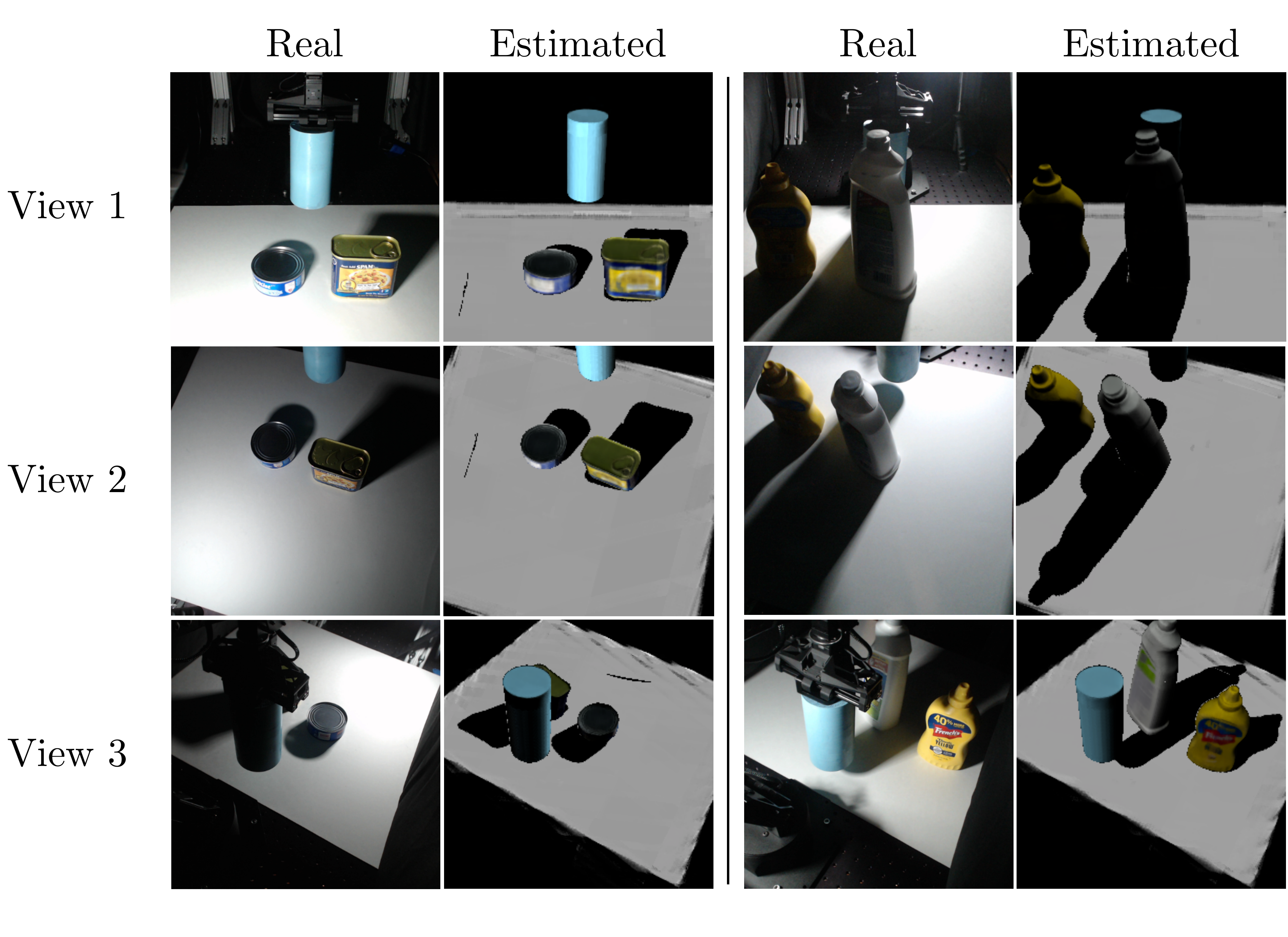}
  \caption{(Top): Lighting estimates produced by inverse parameter estimation from real images. Shadows are well-approximated by our method. (Bottom): Visualized object pose estimates. It is challenging to perfectly optimize the orientation of the bleach bottle due to local minima in appearance-based objectives.
  }
  \vspace{-2em}
  \label{fig:real_estimate_demo}
\end{figure}

Finally, we demonstrate our lighting and pose estimation pipelines with stimuli from a real scene. We collect images in real-world harsh lighting settings created by positioning a single spotlight and blocking external light from the scene. 

For light optimization, we optimize not only the light source position as in the simulated experiments, but also the light distance, ``look at'' direction, and intensity. In the top of Figure~\ref{fig:real_estimate_demo}, we show qualitative examples of the light poses optimized by our method. We see that our method is able to recover a reasonable estimate of the lighting and creates shadows similar to those in the scene.

For real-world object pose optimization, we initialize pose estimates for the optimization process using a pre-trained PoseCNN~\cite{xiang2018posecnn}. This provides a helpful but imperfect initialization that our inverse parameter estimation procedure finetunes. We compute the loss as an intersection-over-union between object masks and masks from the compositional KiloOSF rendering process. In the bottom of Figure~\ref{fig:real_estimate_demo}, we show examples of real observations along with rendered scenes containing objects with poses estimated by our method. The lighting parameters are also determined using our method. Our method is able to reconstruct challenging scenes with reasonable fidelity.

\section{Conclusion}
\label{sec:concl}
We have presented a method that uses object-centric dynamics modeling for robotic manipulation in scenes with unseen object configurations and harsh lighting. We have demonstrated that by leveraging object-centric neural scattering functions, we can invert the rendering procedure to determine object poses and lighting information. This makes our method adaptable to harsh lighting settings and enables us to combine it with a learned neural network dynamics model for use in model-predictive control on long horizon control tasks. We show through experiments that our method achieves better reconstruction in harsh lighting scenarios than previous implicit modeling strategies and improved long horizon prediction and model-predictive control performance compared to video prediction models. 

\paragraph{Limitations.} Currently, our method requires separate KiloOSF models to be trained for each object to be manipulated. Although advancements in training NeRFs efficiently may accelerate this process, this may be time-consuming for large numbers of real objects. We use a simple lighting model; more complex lighting interactions may help model in-the-wild conditions. Additionally, in this work we only consider rigid object manipulation.

\paragraph{Acknowledgments.} This work is in part supported by NSF RI \#2211258, ONR MURI N00014-22-1-2740, AFOSR YIP FA9550-23-1-0127, the Stanford Institute for Human-Centered AI (HAI), the Toyota Research Institute (TRI), Amazon, Ford, Google, and Qualcomm. ST is supported by NSF GRFP Grant No. DGE-1656518. 

{\small
\bibliographystyle{ieee_fullname}
\bibliography{egbib}
}

\end{document}